\documentclass{article}
\usepackage{ijcai22}

\usepackage{times}
\usepackage{soul}
\usepackage{url}
\usepackage[hidelinks]{hyperref}
\usepackage[utf8]{inputenc}
\usepackage[small]{caption}
\usepackage{graphicx}
\usepackage{amsmath}
\usepackage{amsthm}
\usepackage{booktabs}
\usepackage{times}
\usepackage{latexsym}

\usepackage{microtype}
\usepackage{rotating}
\usepackage{natbib}

\usepackage{tabularx}
\usepackage{array, caption, multirow, tabularx,  ragged2e,  booktabs}
\usepackage{multirow}
\usepackage{multicol}
\usepackage{tabularx}
\usepackage[linesnumbered,ruled,vlined]{algorithm2e}

\SetCommentSty{mycommfont}
\usepackage{tikzsymbols}
\usepackage{pifont}
\usepackage[scriptsize]{subfigure}
\usepackage{amsmath,amsfonts,amssymb,amscd,amsthm,xspace}
\SetKwInput{KwInput}{Input}                
\SetKwInput{KwOutput}{Output}              

\usepackage{xcolor, soul}

\sethlcolor{pink}

\usepackage{algpseudocode}
\title{Real-time Caller Intent Detection In Human-Human Customer Support Spoken Conversations}

\author{
    Author Name
    \affiliations
    Affiliation
    \emails
    pcchair@ijcai-22.org
}

\author{
Mrinal Rawat$^1$
\and
Victor Barres$^1$
\affiliations
$^1$Uniphore Software Systems Inc.\\
\emails
\{mrinal.rawat, victor\}@uniphore.com
}

\begin{document}

\maketitle
\begin{abstract}
Agent assistance during human-human customer support spoken interactions requires triggering workflows based on the caller's intent (reason for call). Timeliness of prediction is essential for a good user experience. The goal is for a system to detect the caller's intent at the time the agent would have been able to detect it (Intent Boundary). Some approaches focus on predicting the output offline, i.e. once the full spoken input (e.g. the whole conversational turn) has been processed by the ASR system. This introduces an undesirable latency in the prediction each time the intent could have been detected earlier in the turn. Recent work on voice assistants has used incremental real-time predictions at a word-by-word level to detect intent before the end of a command. Human-directed and machine-directed speech however have very different characteristics. In this work, we propose to apply a method developed in the context of voice-assistant to the problem of online real time caller's intent detection in human-human spoken interactions. We use a dual architecture in which two LSTMs are jointly trained: one predicting the Intent Boundary (IB) and then other predicting the intent class at the IB. We conduct our experiments on our private dataset comprising transcripts of human-human telephone conversations from the telecom customer support domain. We report results analyzing both the accuracy of our system as well as the impact of different architectures on the trade off between overall accuracy and prediction latency.
\end{abstract}

\section{Introduction}
\label{intro}

Conversational AI finds one a key application in reducing the time-to-resolution of customer support calls. It does so by assisting the support agents during a call as well as in their after call work \cite{9117d4d30b3e4e328cd0e3f8ccc0fa2f}. If after call work usually focuses on call disposition, assistance during the call consists in triggering actions at the right time. Such actions can encompass surfacing relevant information, extracting values from the ongoing call, suggest strategies, etc. In this work, we focus on an application of conversational AI to agent assistance during the call. 

A typical contact center solution provides agents with predefined workflows to follow depending on the caller's intent. One of the first tasks of the agent is therefore to recognize the reason for the call and select the appropriate workflow.  We will focus in this work on the goal of detecting the caller's intent in real time. 

Although some systems are designed to operate directly on the speech signal, we focus here on NLU solutions processing the output generated by an automated speech recognition (ASR) system. ASR systems convert audio streams into output word sequences. The frequency at which the output is generated vary. In this work however, without loss of generality, we abstract away the considerations related to ASR output frequencies by assuming that transcriptions will be available one word at a time.

Previous attempts \cite{DBLP:journals/corr/abs-1902-10909, Haihong2019ANB, Qin2019ASF} approached these problems using machine learning and deep learning based techniques in which predictions are carried out at the end of each utterance or turn. An obvious limitation is that the AI system has to wait until the entire word sequence is available before triggering actions, even when enough cues are available early on for a human to reliably trigger the same action. Consider an example, \texttt{i'm calling because i wanted to change my phone number but aah if this call gets disconnected with you call me back}, it is evident that as soon as the word \texttt{number} appears, the agent can predict the caller's intent, but a model working on entire turns would have to wait till the word \texttt{back} is available to generate its prediction. 

A lag between machine prediction and human prediction can lead to increased cognitive load and overall poor user experience. This could in turn make the use of conversational AI systems counterproductive. 

We propose a novel solution where we process the ASR output incrementally (one word at a time) and generate a prediction at each step. Our approach comprises of two step process 1) Intent Boundary (IB) detection: predicts whether the input word is the end of the intent statement 2) Intent detection: predicts the intent. Both these tasks are trained in a multi-task setting. Further, we propose the use of lookahead i.e. incorporating future right context to assist the intent boundary detection. We also study the model performance without the intent boundary task. Finally, we propose two metrics Mean Turn Difference (MTD) and Mean Position Difference (MPD) that aims to understand the effect of the Intent Boundary task in the early prediction.

\section{Related Work}

Most of the previous research works mainly focus on intent detection using a full input sequence. \cite{7078634, Zhang2016AJM,inproceedings} proposes a method that aims to train the NLU tasks such as intent detection, slot filling, etc. jointly using RNN and attention based approaches. Recently, some works  \cite{Qin2019ASF} have also tried to use transformers based architectures such as BERT\cite{DBLP:journals/corr/abs-1810-04805}. \cite{wu-etal-2020-slotrefine} proposes a two-pass refine mechanism to handle uncoordinated slots problem. They claim to have better performance than CRF based architectures.

In recent years, the demand for real-time applications has increased tremendously. This includes incremental processing of the sequence outputting the prediction at each time step. \cite{inproceedings1} first proposes a way that requires the input to be continually
processed based on incremental input. Recently, \cite{DBLP:conf/slt/Shivakumar0GN21} proposed a model that jointly trains two tasks 1) Intent Detection and 2) End-of-sentence(EOS) boundary using LSTM and generates the output at each time step. To imitate the real-time processing the authors randomly concatenate the sentences from the ATIS dataset and consider the last token as the EOS. \cite{ma-etal-2019-stacl} also uses a similar approach where they use an incremental encoder by limiting each source word to attend to its previous input and recompute the representation.
\cite{DBLP:journals/corr/abs-1909-13790} also process the input incrementally while boosting the performance in noisy datasets using data augmentation techniques.
A thorough investigation of the use of non-incremental encoders such as transformers for the incremental systems was conducted by 
\cite{madureira-schlangen-2020-incremental}. \cite{DBLP:conf/emnlp/KahardiprajaMS21} proposed an approach that uses a restart-incrementally approach along with the linear transformers \cite{katharopoulos_et_al_2020} that reduces the overall quadratic complexity to linear. Our approach process the input sequence incrementally using the LSTM network by incorporating the lookahead based intent detection.

\section{Dataset}
\label{section:dataset}

In this work, we use our organization's private dataset which comprises manually annotated transcripts of human-human spoken telephone conversation from the telecom customer support domain. Transcripts were generated by our in-house Kaldi-based ASR system, fine-tuned for this particular domain. Speaker diarization and recognition was not required since agent's and customer's speech was recorded on separate channels. Each transcript defines a sequence of turns 
(single speech sequence until the other speaker starts speaking). Each turn is and each turn is further broken into utterances based on inter lexical pause duration. The time offsets for each word is provided by the the ASR engine. We use this information to analyze the impact of real-time prediction as compared to offline prediction. Table \ref{table:example} describes an example transcript where the highlighted text represents the span of the intent.

\begin{table}[h!]
\centering
\begin{tabular}{|l|r|} \hline
\textbf{Agent} & \textbf{Customer} \\ \hline
you're welcome & \\ to XXX my &\\ 
name is XXX great & \\
how can i help & \\ you today &   \\\hline
& um yeah uh \\
& calling to \hl{i} \\ & \hl{guess uh remove} \\
& \hl{my grandma's} \\ & \hl{phone number} \\ & from her account   \\\hline
\vdots & \vdots \\ \hline
\end{tabular}
\caption{Example}
\label{table:example}
\end{table}

We collect 1544 transcripts and asked annotators to highlight the first text span that would allow a reasonable agent to identify the caller's intent. We also asked that the intent be  classify into one of 16 different intent domain relevant classes. Although in very few transcripts the caller mentions multiple reasons for contacting customer support, the vast majority of the transcripts contain a single intent. The goal is therefore to identify the first and most often only intent stated by the customer.

For our work, we consider only the turns where the intent is present which comprises of 1917 samples and is split into (80\%) train, (20\%) validation and (20\%) test sets. We provide the exact numbers for each split/class in Table~\ref{table:stats} for clarity. 

\begin{table}[h!]
\centering
\begin{tabular}{|l|r|} \hline
\textbf{Split} & \textbf{Total} \\ \hline
Training & 1526  \\\hline
Validation & 194 \\\hline
Test &  197 \\\hline
\textbf{Total} & \textbf{1917}\\\hline
\end{tabular}
\caption{Dataset Statistics}
\label{table:stats}
\end{table}

Table \ref{table:tr_examp} provides an example of training set which describes the input and output format of both tasks. We set the IB token to 1 and rest of the tokens to O.

\begin{table}[h!]
\centering
\resizebox{0.99\columnwidth}{!}{
\begin{tabular}{|l|c|c|c|c|c|c|c|} \hline
\textbf{Sentence} & I & need & my & account & number & rotor & number .. \\ \hline
\textbf{IB} & O & O & O & O & 1 & O & O .. \\ \hline
\textbf{Intent} & O & O & O & O & AM & O & O .. \\ \hline

\end{tabular}
}
\caption{Training example: AM refers to the `Account Management` intent.}
\label{table:tr_examp}
\end{table}

\begin{figure}[h!]
\centering
\subfigure[Distribution of offset difference]{
\includegraphics[width=.90\columnwidth]{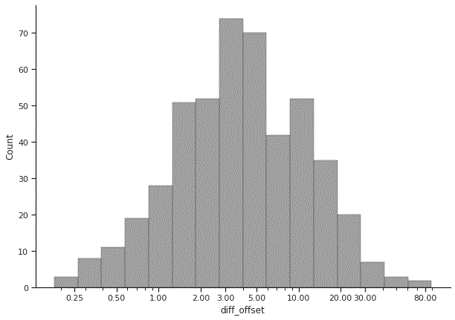}
}
\subfigure[Cumulative Distribution of offset difference]{
\includegraphics[width=.90\columnwidth]{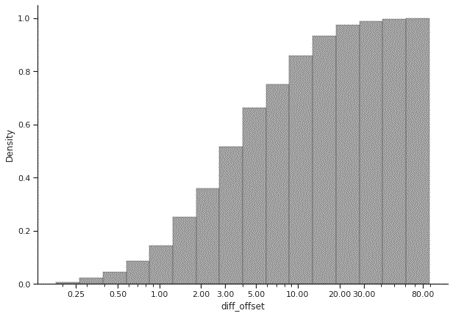}
}
\caption{Distribution of the difference in the offset of intent statement and end of the turn. Here the difference is in \textbf{seconds}}
\label{fig:dist}
\end{figure}

\begin{figure*}[ht!]
\fbox{\includegraphics[trim={4cm 5.9cm 3.5cm 4.8cm},clip,scale=0.87]{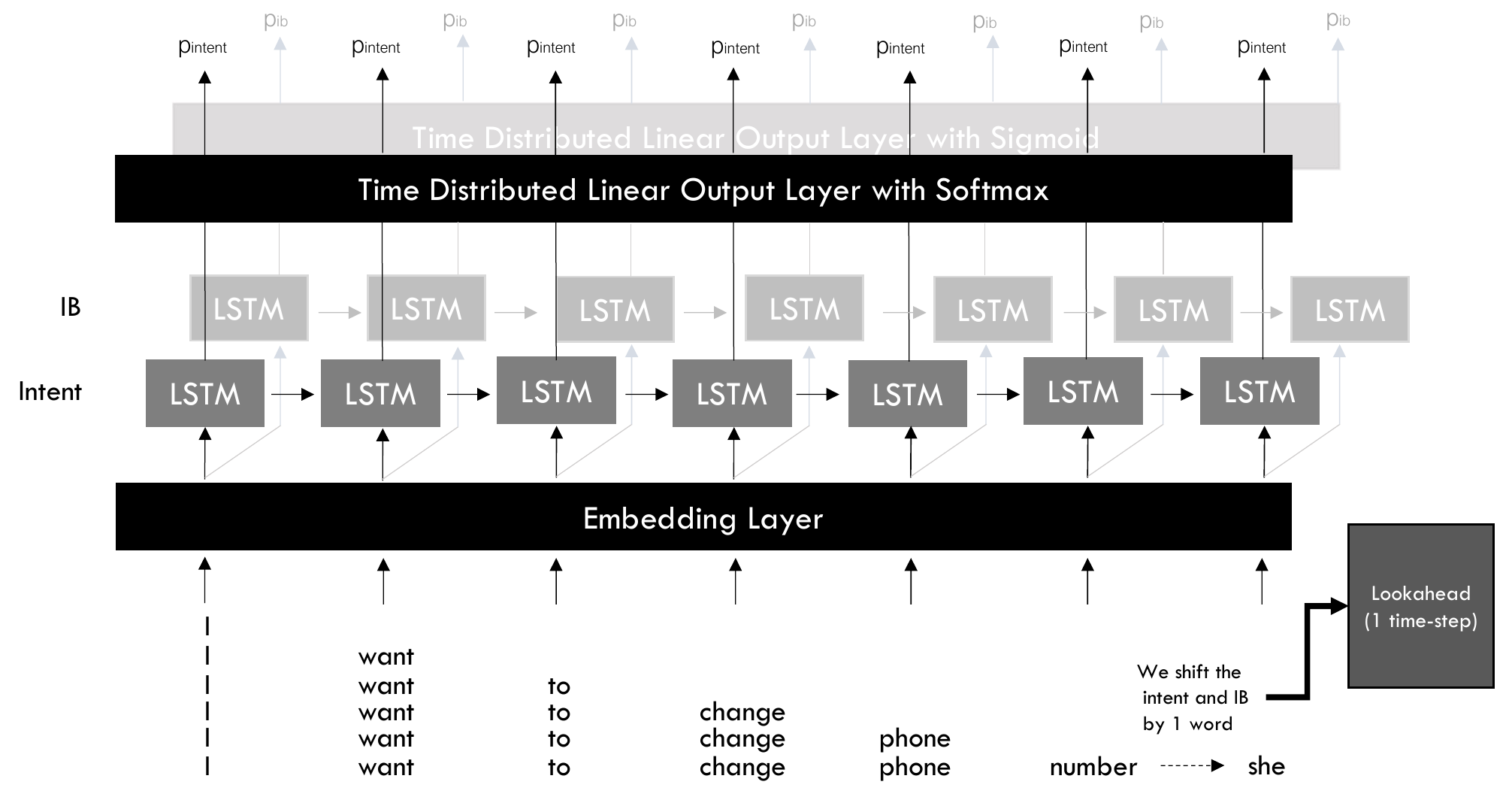}}
\caption{Real-Time Intent Detection Architecture}
\label{fig:fullarchi}
\end{figure*}

Further, we performed an analysis on the dataset to understand the advantages of employing the real-time systems. We discover that \textbf{61.84 \%} of the conversations had the intent at the end of the turn which implies that there is no additional benefit of using a real-time system. In the remaining \textbf{38.16 \%} transcripts, we plot the distribution of the difference in the offset of the intent statement i.e. last uttered word of the highlighted intent statement and the end of the turn (See Figure \ref{fig:dist}). This helps us to measure the average time saved when our approach is used.  This clearly demonstrates that in \textbf{80\%} of the remaining calls the time duration between the statement of intent and the end of the turn is less than \textbf{10 seconds}.

\section{Approach}
\label{approach}

In this section, we first discuss the methodology used for the offline intent detection and then describe the real-time intent detection approach (See Figure \ref{fig:fullarchi}).

\subsection{Offline Intent Detection (Turn Level})
\label{offline}

We use the standard LSTM architecture used in supervised classification problem to detect the intent in an offline way. We use an entire input sequence i.e. turn consisting of sequence of words $w = (w_1, w_2, ..., w_T)$ and pass it to an embedding layer. The output from the embedding layer is then passed into a BiLSTM network which finally assigns an intent class $c$ from a pre-defined set of intents, such that :
     \begin{equation}
     \begin{split}
     \hat{c} = \underset{c}{arg \ max} \ \ P(c|w)
        \end{split}
    \label{eq:offline_eq}
    \end{equation}

\subsection{Real-time Intent Detection}
We use a similar technique proposed by \cite{DBLP:conf/slt/Shivakumar0GN21} for our dataset. The approach comprises two tasks 1) Intent Boundary Detection and 2) Intent Detection which is described below. Both the tasks use similar architecture with slight modifications. To train the network in a real-time setting, we pass the input sequence consisting of words $w = (w_1, w_2, ..., w_T)$ incrementally, i.e. one word at a time. The input word $w_t$ is passed to an embedding layer followed by a uni-directional LSTM. Finally, we pass it to a dense layer and generate the output at each time step. Since the processing is performed incrementally, we store the hidden states generated by LSTM and forward them to the next word $w_{t+1}$. 

\begin{itemize}
    \item \textbf{Intent Boundary (IB) Detection :} This task detects whether the given input word is the intent boundary or not. Since there is no notion of boundary or where to stop, thus we independently train the network with this objective. We use the sigmoid activation function at the output since the task is the binary-classification problem while keeping the rest of the architecture the same. We use the binary focal loss described in Equation \ref{eq:loss_ib} as there is a huge class imbalance in the IB task.
    
     \begin{equation}
     \begin{split}
    L_{IB} & = - \frac{\alpha}{T}\sum_{t=1}^{T} (y_t (1-p_t)^\gamma log(p_t) +  \\
        & (1-y_t)(p_t)^\gamma log(1-p_t))
        \end{split}
    \label{eq:loss_ib}
    \end{equation}
  
  where t is the $t^{th}$ time step, $\alpha$ and $\gamma$ are the hyper-parameters of the focal loss \cite{DBLP:journals/corr/abs-1708-02002}, $y_t$ is the ground truth, and $p_t$ is the output from sigmoid function.
    
    \item \textbf{Intent Detection :} We employ a similar method described above where we predict the intent for each word. At the final level, we use the softmax function since intent detection is a multi-class classification problem. We want to highlight that the intent is the same for the entire sequence, hence we mask the loss function with an indicator function and only compute loss at the Intent boundary. The loss function used here is :
      
      \begin{equation}
            L_{INT} = -\sum_{t=1}^{T} I_{IB} \sum_{c=1}^{C} y_{o,c} \ log \ 
            \hat{y}_{o,c} 
    \label{eq:loss_int}
    \end{equation}
    
    where $T$ is the length of the sequence, $I_{IB}$ is equal to 1 at intent boundary , $y_{o,c}$ is the actual ground truth if o belongs to class c, $\hat{y}_{o,c}$ is the output from the network.
    
\end{itemize}

\subsubsection{Multi-Task Training Objective}
We train both the Intent Boundary (IB) and Intent detection jointly in a multi-task setting where we share an embedding layer and create separate task-specific LSTMs. We train the network with the following objective:

      \begin{equation}
            Loss = \beta L_{IB}  +  (1-\beta) L_{INT}
    \label{eq:loss_final}
    \end{equation}

where $\beta$ is the hyperparmater used to assign weightage to the tasks. We set the $\beta$ to 0.5 after performing a detailed empirical  evaluation.
    
\subsubsection{Lookahead based Intent Detection}

Since the intent detection task is highly dependent on IB detection, improving boundary detection task leads to better performance. We analyze the impact that delaying prediction has on IB detection performance. To do so, we experiment by adding the option of a lookahead i.e. including the future right context where we shift the IB token and the intent token by $k$ words. We evaluated lookahead of one, two and three and concluded that a lookahead of more than one word didn't correlate with better performance. Hence we set the lookahead to 1 (See Table \ref{table:lookahead_ex}). We would also like to note that in the cases where the IB is at the end of the turn, we compared appending for lookahead either random word or the first token of the next turn after the end of the turn. We found that the performance was identical, hence we use the randomly selected token for lookahead. We use the same methodology as described in the above section on the modified dataset.

\begin{table}[h!]
\centering
\resizebox{0.99\columnwidth}{!}{
\begin{tabular}{|l|c|c|c|c|c|c|c|} \hline
\textbf{Sentence} & remove & my & grandma's & phone & number & she & got .. \\ \hline
\textbf{IB} & O & O & O & O & 1 & O & O.. \\ \hline
\textbf{Intent} & O & O & O & O & AM & O & O .. \\ \hline
& & & & & $\rightarrow$& & \\ \hline

\textbf{IB (lookahead)} & O & O & O & O & O & 1 & O.. \\ \hline
\textbf{Intent (lookahead)} & O & O & O & O & O & AM & O .. \\ \hline

\end{tabular}
}
\caption{Example sentence of lookahead based approach}
\label{table:lookahead_ex}
\end{table}

\subsubsection{Context based Detection}

We further investigate the use of context in the intent detection task. We hypothesize that the previous context, i.e previous $n$ turns, can assist the network in improving the performance of the tasks. For instance, based on our analysis we found that intent is usually present in the turn following the agent's greeting message e.g. 

\textit{
AGENT : Hello my name is James and how can i help you today \\ 
CUSTOMER: Hi uh i wanted to change my phone number.}

To encode the context, we take the previous $3$\footnote{Selected with empirical evaluation and manual analysis after trying 2, 3, 4} turns and concatenate them and pass them to an LSTM layer. We further initialize the network for the IB and intent detection with the hidden states generated from the context LSTM.  We note that the embedding layer is shared across the tasks and the rest of the network is optimized using the method above.

\begin{table*}[h!]
  \begin{center}
  \resizebox{2.1\columnwidth}{!}{
    \begin{tabular}{|l|c|c|r|r|r|r|r|r|r|r|r|r|} 
    \hline
     & \multicolumn{3}{c|}{Intent Boundary} & \multicolumn{3}{c|}{Intent @ OB} & \multicolumn{3}{c|}{Intent @ PB} & \multicolumn{3}{c|}{Intent}  \\ 
        \cline{2-13}
\textsubscript & P & R & F1 & P & R & F1 & P & R & F1 & P & R & F1\\
      \hline
      Offline (Turn Level) & - & - & - & - & - & - & - &  - & - & \textbf{0.4271} & 0.3120 & 0.3463  \\
       \hline
      \multicolumn{13}{|c|}{With Intent Boundary} \\
       \hline
      Baseline & \textbf{0.4671} & 0.4702 & 0.4686 & 0.8285 & 0.8122 &  0.8205 & \textbf{0.3521} & \textbf{0.4745} & \textbf{0.389} & - & - & -\\
      Baseline + Lookahead & 0.430 & 0.572 & 0.491 & 0.8259 & 0.8183 & 0.8221 & 0.255 & 0.4136 & 0.3032 & - & - & - \\
      Baseline + Context & 0.4659 & \textbf{0.623} & \textbf{0.5328} & \textbf{0.8551} & \textbf{0.8375} & \textbf{0.8364} & 0.2194 & 0.4152 & 0.2814 & - & - & - \\ 
      \hline 
     \multicolumn{13}{|c|}{Without Intent Boundary} \\
         \hline
      Intent + Lookahead & - & - & - & -& - & - & - & - & - & 0.399 & \textbf{0.5577} & \textbf{0.4532}\\ 
          \hline
    \end{tabular}
    }
    \caption{Evaluation of Intent Boundary (IB), Intent, Intent @ Oracle Boundary (OB), and Intent @ Predicted Boundary (PB) on the test set using precision (P), recall (R) and F1-score (F1).}

    \label{tab:Evaluation-test1}
  \end{center}
\end{table*}

\section{Experimental Setup}
\label{expsetup}

We train both the tasks i.e 1)Intent Boundary Detection and 2) Intent Detection in a multi-task learning framework. The embedding layer is initialized randomly with the size 300 and is shared across both tasks. We use task-specific separate LSTMs with hidden dimensions of 128 and the number of layers as 2 for both tasks. We trained the network for 30 epochs using Adam optimizer with a learning rate of 0.001 and drop out as 0.25. We use the focal loss to optimize the IB task and set its hyperparameters $\alpha$ to 1 and $\gamma$ to 8. We ran experiments multiple times to ensure model convergence. All models are implemented in Pytorch \cite{NEURIPS2019_9015}.

\subsection{Evaluation Metrics}
\label{metrics}

We perform evaluation at two levels:

\textbf{Model Evaluation:}

We evaluate both the tasks along with the combined performance by looking at precision, recall and f1-score computed on a test set composed only of turns containing an intent.

\textbf{Real Time Evaluation:}
Since the goal is for the model to operate on conversations in real time,
we analyze the intent detection performance on temporally unfolding conversations. Given a transcript, the customer turns are passed to the system in order, and each turn is processed one word at time. The hidden states of the network are reset before a new turn starts.  We stop when an intent boundary and its associated intent class $c$ is detected (See Algorithm \ref{algo}). For the lookahead based approach, we also add a random token at the end of the turn. We evaluate the performance of the real-time processing using metrics defined below:

\begin{itemize}
    \item \textbf{Accuracy:} It is computed by comparing the first predicted intent and the ground truth.
    
    \item \textbf{Accuracy at right turn:} It is computed by comparing the first predicted intent and the turn position with the ground truth intent and turn position.

    \item \textbf{Accuracy at right position:} It is computed by comparing the first predicted intent, turn position, and intent position within the turn with ground truth intent, turn position, and intent word position.
    
   \item  \textbf{Mean Turn Difference (MTD):} The average difference between the ground truth intent turn position and the predicted intent turn position. 
                \begin{equation}
                    MTD = \frac{1}{N} \sum_{n=1}^{N} t_{n} - \hat{t}_{n}
                \end{equation}
    where N is the number of samples in the evaluation set, $t$ is the actual turn position and $\hat{t}$ is the predicted turn position.
    
   \item \textbf{Mean Position Difference (MPD):} The average difference between the ground truth intent word position and the predicted intent word position.
   \begin{equation}
                    MPD = \frac{1}{N} \sum_{n=1}^{N} p_{n} - \hat{p}_{n}
                \end{equation}
                
         where N is the number of samples in the evaluation set, $p$ is the actual word position and $\hat{p}$ is the predicted word position.
\end{itemize}

\begin{algorithm}[]
\DontPrintSemicolon
  \KwInput{transcripts $t$, model, true\_intents}

\SetKwFunction{SearchF}{RealTimeEvaluation}
  \SetKwProg{Fn}{Function}{:}{}

  $pred\_intents \gets \emptyset$;
  
  \ForEach{$t_i \in t$}{%
    
    \ForEach{$turn \in t_i$}{%
    $previous\_intent\_state \gets \emptyset$;\\
    $previous\_ib\_state \gets \emptyset$;\\
            \If{$turn['speaker']$ == 'customer'}{
                \ForEach{$token \in turn$}{
                    $o\_ib$, $score$, $previous\_ib\_state$ $\gets$ model.predict\_ib($token$,
                    $previous\_ib\_state$)
                    
                    \If{$ o\_ib = 1$ \& $score > T$}{
                    $o\_int$, $previous\_intent\_state$ $\gets$ model.predict\_intent(token,\\ previous\_intent\_state)
                    
                    $pred\_intents.add$($o\_int$);\\
                    break \{turn loop\};
                    }
                }
            }
        
    }
        
  }
  $Get\_Metrics(true\_intents, pred\_intents)$\\

\caption{Pseudo-code for the real-time evaluation}
\label{algo}
\end{algorithm}

\begin{table*}[h!]
  \begin{center}
    \begin{tabular}{|l|c|c|r|r|r|} 
    \hline
     & \multicolumn{5}{c|}{Real-Time Evaluation} \\ 
        \cline{2-6}
\textsubscript & Acc & Acc @ RT & Acc @ RP & MTD & MPD\\
      \hline
      Offline (Turn Level) & 0.7697 & 0.7236 & - & -0.78 & - \\
       \hline
      \multicolumn{6}{|c|}{With Intent Boundary} \\
       \hline
      Baseline  & 0.7763 & \textbf{0.7631} & 0.3618 & -0.59 & -13.56 \\
      Baseline + Lookahead  & 0.7894 & 0.7236 & 0.4539 & \textbf{ -0.88} & \textbf{-19.56}  \\
      Baseline + Context  & 0.75& 0.6578 &0.4539 &-0.344 & -7.76 \\ 
      \hline 
     \multicolumn{6}{|c|}{Without Intent Boundary} \\
         \hline
      Intent + Lookahead  & \textbf{0.8355} & 0.6842 & \textbf{0.4934} & 2.37 & 44.66 \\ 
          \hline
    \end{tabular}
    \caption{Evaluation on the test set by simulating the conversation in real-time using accuracy (Acc), accuracy at right turn (Acc @ RT), accuracy at right position (Acc @ RP), mean turn difference (MTD), and mean position difference (MPD)}

    \label{tab:Evaluation-test-offline}
  \end{center}
\end{table*}

\section{Results \& Discussion}
\label{results}

We evaluate our approach using the metrics defined in Section \ref{metrics} The overall results are described in two parts 1) Evaluation of the test dataset created using the intent statements from the transcripts 2) Evaluation of the test data by simulating the actual conversation in a real time environment.

The results for part 1 are illustrated in Table \ref{tab:Evaluation-test1} using the methods described in Section \ref{approach}.  We break the results task wise where we describe the precision, recall, and F1 score for Intent Boundary (IB) task, Intent @ oracle boundary and Intent @ Predicted boundary. We can clearly see that incorporating lookahead yields a better F1-score than the baseline method in the IB task since we also provide the future context to the model. Further, we see that adding context yields the best performance in the Intent Boundary task and Intent @ OB task, but has the worst F1 score of intent @ PB which is the metric we desire for better performance when used in real time setting.

We also investigate the model performance by excluding the Intent Boundary objective. In contrast to the method described in Section \ref{approach}, we compute the loss at each time step since we do not train with explicit intent boundary in this case. We achieve an F1-score of 0.4532 which is computed based on $c$ classes while excluding the $O$ class. Similar to the previous method, we also incorporate the lookahead where we shift the intent towards the right. Finally, we report the results on the offline methodology described in Section \ref{offline} where we classify the entire turn into one of the 
$c$ classes.

Further, we see that the addition of context did not improve the performance proving our initial hypothesis wrong. We achieve an accuracy of \textbf{75 \%} which is the worst amongst all. One of the reasons could be the long sequences due to the addition of context which negatively affected the LSTM performance. However, we want to investigate this in detail by encoding the context using the more advanced methods and comparing it with other datasets as well.

Table \ref{tab:Evaluation-test-offline} illustrates the results where we perform the online i.e. real-time evaluation. For evaluation, we use the same test dataset to compare with previous results and avoid data leakage issues. Similar to the previous results, incorporating lookahead yields better performance. We achieve an accuracy of \textbf{83.55\%} without the IB task outperforming the rest of the models.

\begin{figure}[h!]
\centering
\subfigure[With IB]{
\includegraphics[width=.65\columnwidth]{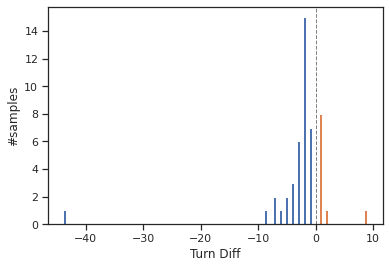}
}
\subfigure[Without IB]{
\includegraphics[width=.65\columnwidth]{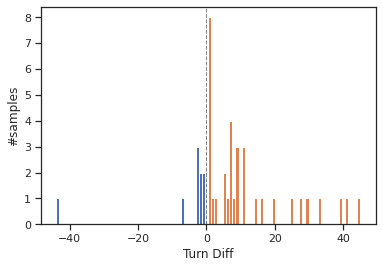}
}
\caption{Distribution of the turn difference in the predicted IB and actual IB with and without using Intent Boundary (IB). Here the difference is in \#turns. Blue color bar represents the early prediction and orange color represents the late prediction}
\label{fig:word_dist}
\vspace{-0.3cm}
\end{figure}

We see a consistent performance in the accuracy at the right turn and accuracy at the right position as well. We also describe the results using the Mean turn difference (MTD) and Mean Position Difference (MPD) which essentially explains the early or late prediction of the intents. Table \ref{tab:Evaluation-test-offline} demonstrates that the inclusion of the intent boundary task helps in predicting the intents much before the actual intent boundary.  Our best model can predict \textbf{0.88 turns} and \textbf{19.56 words} earlier reducing the overall latency. Further, we see that without the explicit IB task, there is a delay of \textbf{2.37 turns} in the prediction even though we achieve the best accuracy of 83 \%. We also plot the distribution of the predicted turn difference w.r.t to the actual turn. Figure \ref{fig:word_dist} clearly demonstrates the importance of including IB for the early prediction. This explains the trade-off between the system with the overall higher accuracy and the system that reduces the latency.

\section{Conclusion and Future Work}
In this work, we present a systematic method to detect intent in real-time in human-human customer support conversations. We propose a dual architecture approach to detect intent and intent boundary by incorporating the lookahead. We also investigate various methodologies such as the adding context, excluding the IB task, etc. We discuss the results on our private dataset comprised of text transcripts of human-human conversations from the telecom domain. To evaluate the methodology in real-time, we also propose metrics such as Mean turn difference (MTD) and Mean Position difference (MPD). We conclude that the intent boundary task significantly helps in reducing latencies by predicting the intents earlier. Our method achieves the best overall intent accuracy of \textbf{83.55 \%} when IB task is excluded.

In the future, we would like to use our method to reproduce this study with other datasets and ensure that our method is consistent. Further, we also want to explore more advanced deep learning techniques such as transformers to encode the sequences and context. We also want to extend this work to predict the entities in real-time.

\bibliographystyle{named}
\bibliography{ijcai22}

\end{document}